\documentclass[conference]{IEEEtran}

\usepackage{multirow}

\usepackage{cite}

\ifCLASSINFOpdf
   \usepackage[pdftex]{graphicx}
\else
\fi

\usepackage[cmex10]{amsmath}
\usepackage{amsfonts}
\usepackage{bm}
\usepackage{xfrac}
\usepackage{mathtools}

\usepackage[linesnumbered, vlined, ruled, commentsnumbered]{algorithm2e}

\usepackage{array}
\usepackage{multicol}
\usepackage{colortbl}

\ifCLASSOPTIONcompsoc
  \usepackage[caption=false,font=normalsize,labelfont=sf,textfont=sf]{subfig}
\else
  \usepackage[caption=false,font=footnotesize]{subfig}
\fi

\usepackage{ragged2e}

\usepackage{balance}

\usepackage{dblfloatfix}

\usepackage{url}
\usepackage[colorlinks=true, citecolor=green, linkcolor=red, urlcolor=blue]{hyperref}
\usepackage[hyperpageref]{backref}

\usepackage{lipsum}

\usepackage{graphicx}
\usepackage[export]{adjustbox}

\usepackage{longtable}
\usepackage{booktabs}
\usepackage{ifthen}

\begin{document}
\bstctlcite{IEEEexample:BSTcontrol}

\title{RepAir: A Framework for Airway Segmentation and Discontinuity Correction in CT}

\author{\IEEEauthorblockN{John M. Oyer$^{1}$, Ali Namvar$^{2}$, Benjamin A. Hoff$^{2}$, Wassim W. Labaki$^{3}$, Ella A. Kazerooni$^{2}$, \\Charles R. Hatt$^{2,4}$, Fernando J. Martinez$^{5}$, MeiLan K. Han$^{3}$, Craig J. Galbán$^{2}$, and Sundaresh Ram$^{2,6}$}
\IEEEauthorblockA{Department of $^{1}$Computer Science, $^{2}$Radiology, $^{3}$Internal Medicine, University of Michigan, Ann Arbor, MI.\\
$^{4}$4D Medical, Inc., Minneapolis, MN; $^{5}$T. H. Chan School of Medicine, University of Massachusetts, Worcester, MA.\\
$^{6}$Department of Radiology \& Imaging Sciences, Biomedical Engineering, Electrical \& Computer Engineering,\\Emory University \& Georgia Institute of Technology, Atlanta, GA.\\
Email: {\tt \href{mailto:johnoyer@umich.edu}{johnoyer@umich.edu},\href{mailto:sundaresh.ram@emory.edu}{sundaresh.ram@emory.edu}}}
}

\maketitle

\begin{abstract}

Accurate airway segmentation from chest computed tomography (CT) scans is essential for quantitative lung analysis, yet manual annotation is impractical and many automated U-Net-based methods yield disconnected components that hinder reliable biomarker extraction. We present \textit{RepAir}, a three-stage framework for robust 3D airway segmentation that combines an nnU-Net-based network with anatomically informed topology correction. The segmentation network produces an initial airway mask, after which a skeleton-based algorithm identifies potential discontinuities and proposes reconnections. A 1D convolutional classifier then determines which candidate links correspond to true anatomical branches versus false or obstructed paths. We evaluate \textit{RepAir} on two distinct datasets: ATM’22, comprising annotated CT scans from predominantly healthy subjects and AeroPath, encompassing annotated scans with severe airway pathology. Across both datasets, \textit{RepAir} outperforms existing 3D U-Net-based approaches such as \textit{Bronchinet} and \textit{NaviAirway} on both voxel-level and topological metrics, and produces more complete and anatomically consistent airway trees while maintaining high segmentation accuracy.

\end{abstract}

\begin{IEEEkeywords}
airway segmentation, image segmentation, convolutional neural networks, deep learning, computational imaging.
\end{IEEEkeywords}

\IEEEpeerreviewmaketitle

\vspace{-3mm}
\section{Introduction}
\label{sec:intro}

Accurate segmentation of human airways from inspiratory chest computed tomography (CT) scans is essential for computing quantitative metrics relevant to lung disease. Airway-based measurements such as wall thickness, lumen diameter, tapering, and total airway count serve as imaging biomarkers in conditions including chronic obstructive pulmonary disease, asthma, bronchiectasis, and cystic fibrosis, and are used to evaluate airway remodeling over time \cite{achenbach2008, mcdonough2011, bodduluri2024, kirby2017, kuo2017, tschirren2005}. More reliable segmentation directly improves the accuracy of these quantitative measures \cite{tschirren2005, achenbach2008}. While manual annotation provides a gold standard, it is prohibitively time-consuming \cite{kuo2017}. Fully automated methods exist, but they often fail to recover peripheral or high-generation airways and can produce disconnected trees, particularly in subjects with severe airway disease, limiting their reliability for clinical or research applications \cite{tang2023, Garcia-Uceda2021}. On CT, airways appear as low-density lumina surrounded by high-density walls, and most airway segmentation methods therefore focus on identifying the lumen, as manual annotations used for training typically delineate the lumen alone.

Early airway segmentation methods relied on region-growing techniques, which expanded candidate regions outward from a seed point until intensity changes indicated airway walls \cite{tschirren2005}. While conceptually simple, these methods faced several challenges: they were prone to leakage when airway walls were indistinct and often failed to capture higher-generation airways where partial-volume effects and image noise become more pronounced \cite{tschirren2005, Garcia-Uceda2021}. Consequently, these approaches struggled to recover the full extent of the airway tree.

\begin{figure*}[!t]
	\centering
    \includegraphics[width= 7in]{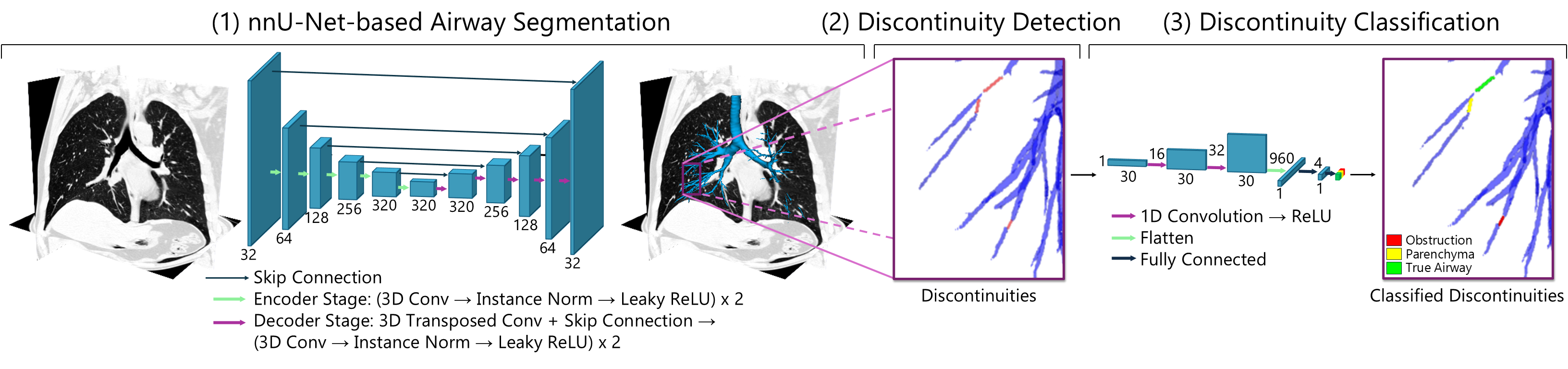}
    \vspace{-4mm}
	\caption{Proposed segmentation pipeline. Teal boxes represent CNN layers. In the nnU-Net-based segmentation network, number of channels is shown below each layer. In the discontinuity classification network, input size is shown below each layer, and number of channels is shown to the left of each layer.}
	\label{fig1}
\vspace{-5mm}
\end{figure*}

Deep learning–based architectures, particularly U-Nets and their variants, have become the standard for airway segmentation \cite{Garcia-Uceda2021, wang2023, tang2023}. U-Nets employ an encoder–decoder structure that captures both local and contextual features. 3D U-Nets with patch-based training efficiently process large CT volumes while preserving local anatomical detail and detect airways across multiple scales and complex topologies. Nevertheless, segmenting small, high-generation airways remains difficult because their diameters approach the spatial resolution limit of CT, reducing contrast relative to image noise. In addition, U-Nets do not inherently guarantee a single connected airway tree, often resulting in disconnected components. Peripheral airways are particularly challenging due to class imbalance between airway and background voxels.

Several recent studies have addressed these challenges through architectural and loss-function innovations. Garcia-Uceda \emph{et al.} \cite{Garcia-Uceda2021} developed \textit{Bronchinet}, a 3D U-Net trained with large cropped patches ($252\times252\times252$) and a soft Dice loss masked to the lungs, allowing improved lumen detection and robust overlap-based optimization. However, its reliance on large patch sizes constrained flexibility and limited generalizations to scans of different resolutions or fields of view. Wang \emph{et al.} \cite{wang2023} introduced \textit{NaviAirway}, another 3D U-Net variant that incorporated dilated convolutions and self-attention to enhance feature extraction, along with a "skeleton Dice loss" designed to preserve topological continuity. They also used a student-teacher framework to leverage unlabeled data for semi-supervised training. Although these methods advanced voxel-level accuracy and improved topological consistency, issues of incomplete connectivity and fragmented small-airway reconstruction persist, particularly in subjects with severe or diffuse airway disease.

Our work differs from previous approaches in two main aspects: (1) we develop a connection algorithm and a classifier to repair broken airway segments, applicable to any segmentation method that does not guarantee a single connected component; and (2) we validate our method on two distinct datasets---one consisting predominantly of healthy cases and another including subjects with severe lung disease. We describe our algorithm in detail and present quantitative comparisons against existing methods.

\section{Methods}
\label{sec:mhtd}

Our proposed model, \textit{RepAir}, is a three-stage framework designed to generate anatomically consistent airway segmentations from chest CT scans (Fig.~\ref{fig1}). The model integrates a robust segmentation network with topological correction and anatomical validation.
\subsection{RepAir Architecture and Processing Pipeline}

The framework begins with a 3D U-Net producing an initial airway mask, followed by a skeleton-based algorithm that detects candidate discontinuities. A 1D convolutional neural network (CNN) then classifies each discontinuity as \textit{true airway}, \textit{parenchyma}, or \textit{airway obstruction}. Together, these stages produce structurally consistent airway reconstructions without compromising voxel-level accuracy. Below, we describe each stage in detail. 

\textbf{(1) 3D nnU-Net-based Segmentation Network:} We adopted nnU-Net \cite{isensee2021} as the backbone architecture for our airway segmentation network. The standard U-Net configuration was used, as alternative configurations, such as the ResNet backbone or "cascade" architectures, did not yield improvements in segmentation performance for our datasets. During training, the input CT volumes were divided into 3D patches of size $128\times160\times112$, allowing efficient processing while preserving sufficient spatial context. The network was trained using a combination of Dice loss and cross-entropy loss, balancing overlap accuracy and voxel-level classification. Optimization was performed using stochastic gradient descent, and data augmentation (including rotations, scaling, noise, blur, intensity adjustments and mirroring) was applied to improve generalization \cite{isensee2021}. The network architecture, including encoder-decoder layout and feature channel dimensions, is illustrated in Fig.~\ref{fig1}.

\textbf{(2) Airway Discontinuity Detection:} Discontinuities in the predicted airway tree were identified using a skeleton-based algorithm, which operates under the assumption that each endpoint of the airway skeleton lying within the true airway tree may correspond to one end of a segmentation discontinuity. For every skeleton endpoint, the algorithm systematically searches for potential connection candidates among nearby skeleton voxels within a specified radius. Candidate connections are gradually added according to a prioritization scheme: short-distance connections are preferred, sharp or anatomically implausible angles are avoided, and only connections that preserve the tree-like structure---i.e., those that do not introduce cycles---are accepted. The exact cutoffs and weights, such as those for angle and distance, were empirically tuned to ensure connections retained realistic airway structure. Once selected, each connection is dilated to match the diameter of the airway segment it joins, ensuring anatomically consistent and continuous airway reconstruction. 

\textbf{(3) Airway Discontinuity Classification:} A 1D CNN classifies each candidate discontinuity as \textit{true airway}, \textit{parenchyma}, or \textit{airway obstruction}. The network input consists of spatially ordered voxels along the centerline of the discontinuity and adjacent airway segments, providing local structural and intensity information for accurate classification. For training, \textit{true airway} samples are drawn from ground-truth airway centerlines, representing healthy airway portions missed by the segmentation network. \textit{Parenchyma} samples are taken from non-airway lung regions to simulate incorrect connections passing through lung tissue, while \textit{airway obstruction} samples are generated from healthy airway segments with added high-density noise to mimic obstructions that appear as discontinuities. Only discontinuities classified as \textit{true airway} are incorporated into the final segmentation, ensuring that the reconstructed airway tree is anatomically consistent and continuous. We used 20,000 1D patches from each of the three classes to train the classifier. Network details are illustrated in Fig.~\ref{fig1}.
\subsection{Datasets}

All models were trained from scratch using a subset of CT scans from the ATM'22 \cite{ZHANG2023} challenge training dataset. Performance was evaluated on a held-out test set from the same dataset and further validated on the AeroPath \cite{stoverud2024} dataset to assess generalization. 

\textbf{ATM'22:} The ATM'22 dataset includes 280 lung CT scans with manual airway annotations by three radiologists. Most scans are from healthy patients, with some from individuals with COVID-19. Axial resolution was $512\times512$, with $157-1125$ slices per scan, slice thickness of $0.50-1.00$ mm, and in-plane resolution of $0.514-0.919$ mm. After discarding one scan due to a size mismatch, the remaining 279 scans were randomly split into 209 for training and 70 for testing, a split used consistently across all models.

\textbf{AeroPath:} The publicly available AeroPath dataset contains 27 CT scans with manual airway annotations of patients exhibiting severe lung pathologies that alter airway morphology. Scans were acquired from patients evaluated for lung cancer and include conditions such as malignant tumors, sarcoidosis, and emphysema. Annotations were supervised and verified by a pulmonologist. Axial resolution ranged from $[487-512] \times [441-512]$, with $241-829$ slices, slice thickness of $0.50-1.25$ mm, and in-plane resolution of $0.68-0.76$ mm.

\subsection{Implementation and Training Details}

The nnU-Net-based \textit{RepAir} segmentation network was trained for 1,000 epochs, where each epoch consisted of 250 mini-batches. Stochastic gradient descent with Nesterov momentum ($\mu = 0.99$) and an initial learning rate of 0.01 was used to optimize the network weights. Instance normalization was applied before each Leaky ReLU (rectified linear unit) activation during training, with global statistics updated according to these instance-specific values as described in \cite{isensee2021}.

\textit{Bronchinet} was originally designed for smaller datasets and loads all images into GPU memory simultaneously, which limited us to using 20 training scans from the full set of 209. The original implementation uses a trachea and large-airway mask to guide the segmentation, but since our goal was full airway tree segmentation, we did not include any such mask and performed one-pass segmentation directly. Although \textit{NaviAirway} supports semi-supervised training with unlabeled data, we trained it using only labeled scans for fair comparison across models.

All models (\textit{RepAir}, \textit{Bronchinet} and \textit{NaviAirway}) were trained from scratch, with randomly initialized weights, and evaluated on an NVIDIA Tesla V100 GPU (16GB of memory). Training for each model was completed in under two days. 

\section{Experiments and Results}
\label{exp_res}

We compared our proposed method, \textit{RepAir}, alongside state-of-the-art 3D U-Net-based airway segmentation methods, namely \textit{Bronchinet} and \textit{NaviAirway}. For all methods, including \textit{RepAir} and the benchmark models, only the largest connected component of the final segmentation prediction was analyzed, following prior evaluations \cite{Garcia-Uceda2021, wang2023, ZHANG2023}, as it represents the anatomically meaningful airway structure. 

\subsection{Runtime Evaluation}

\begin{figure}[!t]
    \vspace{-3mm}
    \centering
    \includegraphics[width=2.8in]{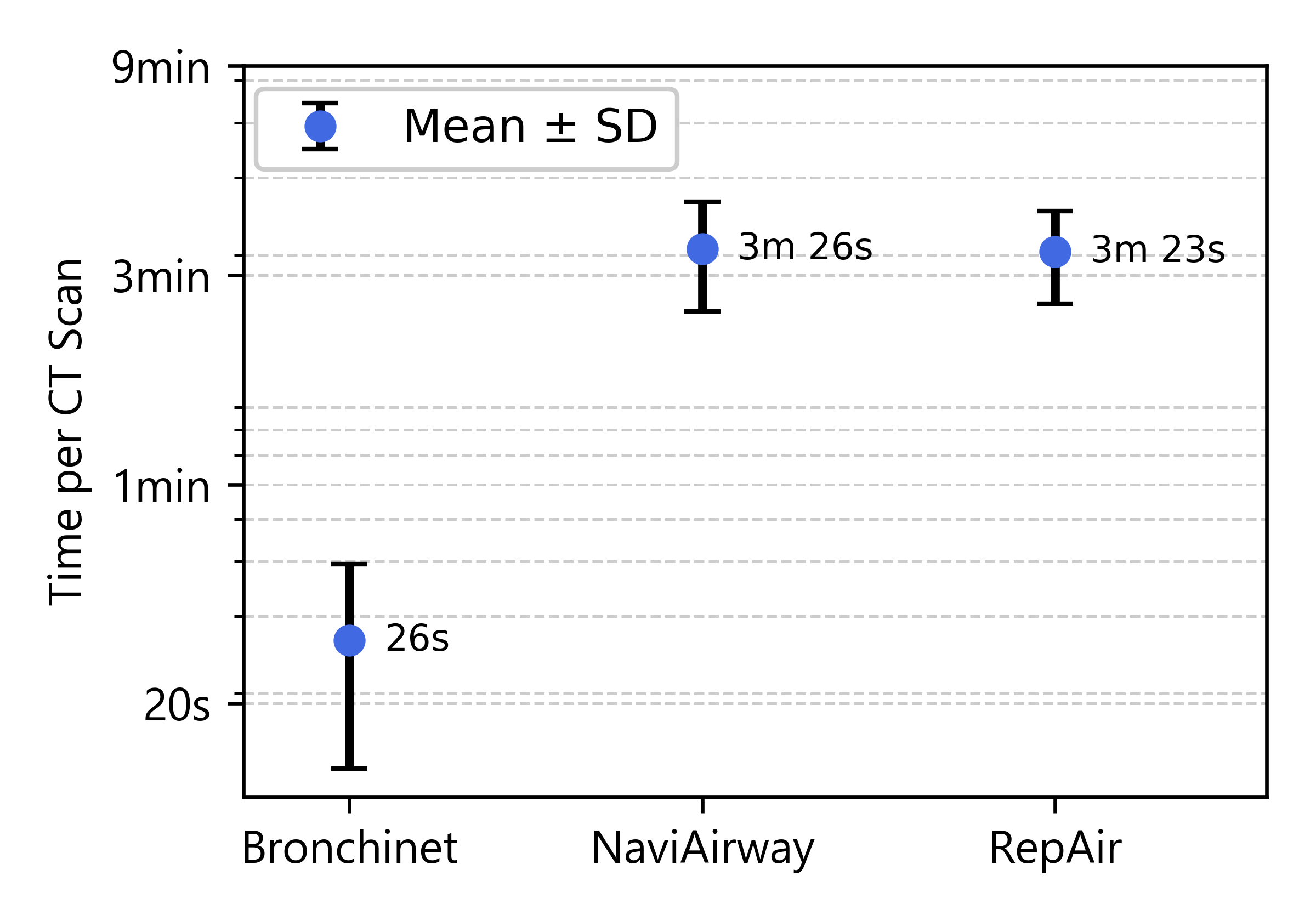}
    \vspace{-4mm}
    \caption{Inference time per CT scan for each method.}
    \vspace{-5mm}
    \label{fig2}
    \end{figure}

Figure~\ref{fig2} compares inference times across methods, showing that \textit{RepAir} runs 2 minutes and 57 seconds slower than \textit{Bronchinet} but 3 seconds faster than \textit{NaviAirway}. The slightly longer runtime relative to \textit{Bronchinet} can be attributed to the greater number of input patches and deeper encoder-decoder structure of the nnU-Net backbone, which allows for improved feature representation at the cost of marginally higher computation. In contrast, its runtime remains comparable to \textit{NaviAirway}, reflecting a similar level of model complexity and memory footprint.

\begin{table*}[!t]
\vspace{-4mm}
\caption{\sc{Segmentation Performance: Mean ± Standard Deviation}}
\label{table1}
  \begin{center}
\vspace{-4mm}
	\renewcommand{\arraystretch}{1.1}
	\begin{tabular}{>{\centering} m{2.8cm} >{\centering} m{2.8cm} >{\centering}m{1.4cm} >{\centering} m{1.4cm} >{\centering} m{1.4cm} >{\centering} m{1.4cm} >{\centering} m{1.8cm} >{\centering} m{1.4cm}}
	\hline
	\rowcolor[gray] {0.8}\textbf{Data} & \textbf{Method} &  \textbf{Dice} & \textbf{TPR} & \textbf{FPR ($\times 10^{\scriptstyle -4}$)} & \textbf{JI} & \textbf{HD} & \textbf{TD} \tabularnewline \hline
    \multirow{5}{*}{\raisebox{+2em}{\shortstack{\textbf{ATM'22 Test Set}\\70 annotated CT scans}}}
	 & RepAir & \textbf{0.93 ± 0.02} & \textbf{0.93 ± 0.04} & \textbf{0.55 ± 0.41} & \textbf{0.88 ± 0.04} & \textbf{42.75 ± 20.53} & \textbf{0.92 ± 0.06} \tabularnewline 
    & Bronchinet & 0.92 ± 0.03 & 0.92 ± 0.05 & 2.38 ± 0.88 & 0.85 ± 0.05 & 44.61 ± 18.32 & 0.82 ± 0.08 \tabularnewline
	& NaviAirway & 0.80 ± 0.05 & 0.83 ± 0.07 & 2.04 ± 0.69 & 0.67 ± 0.07 & 69.96 ± 16.34 & 0.47 ± 0.10 \tabularnewline
	\hline
    \multirow{5}{*}{\raisebox{+2em}{\shortstack{\textbf{AeroPath}\\27 annotated CT scans}}}
	 & RepAir & \textbf{0.88 ± 0.03} & \textbf{0.89 ± 0.08} & 1.73 ± 1.09 & \textbf{0.78 ± 0.05} & \textbf{50.15 ± 17.22} & \textbf{0.90 ± 0.07} \tabularnewline
    & Bronchinet & 0.62 ± 0.21 & 0.51 ± 0.24 & \textbf{1.29 ± 1.42} & 0.48 ± 0.21 & 173.92 ± 64.57 & 0.25 ± 0.21 \tabularnewline
	& NaviAirway & 0.84 ± 0.05 & 0.88 ± 0.08 & 2.46 ± 1.20 & 0.73 ± 0.08 & 79.86 ± 34.95 & 0.63 ± 0.17 \tabularnewline
           \hline
	\end{tabular}
   \end{center}
\vspace{-6mm}
\end{table*}

\subsection{Performance Evaluation}

We evaluated all methods using Dice coefficient (Dice), true positive rate (TPR), false positive rate (FPR), Jaccard index (JI), Hausdorff distance (HD) \cite{sundaresh2020}, and tree detected rate (TD) \cite{stoverud2024,ZHANG2023, Garcia-Uceda2021}. Let $TP$, $FP$, $FN$, and $TN$ denote the numbers of true positive, false positive, false negative, and true negative voxels, respectively.

The Dice coefficient measures overlap between the predicted and the ground-truth airways, defined as: $$\text{Dice} = \frac{2\cdot TP}{2\cdot TP+FP+FN}$$
TPR (recall, sensitivity) and FPR (equal to $1-$specificity) are defined as: $$\text{TPR} = \frac{TP}{TP+FN}; \quad \text{FPR} = \frac{FP}{FP+TN}$$
The Jaccard Index (JI) or Intersection-over-Union, is given by: $$\text{JI}=\frac{TP}{TP+FP+FN}$$
The Hausdorff distance (HD) quantifies the largest Euclidean distance from any boundary voxel in one airway segmentation (predicted or ground-truth) to the nearest boundary voxel in the other, capturing boundary-level discrepancies \cite{sundaresh2020}. 

TD quantifies how much of the ground truth airway tree length is recovered by the prediction and is defined as the ratio of prediction centerline length to ground truth centerline length \cite{stoverud2024,ZHANG2023, Garcia-Uceda2021}.

Together, these metrics assess both voxel-level segmentation accuracy (Dice, TPR, FPR, and JI) and airway topology continuity (TD, HD). The best segmentation algorithm is therefore one that maximizes the Dice coefficient, TPR, JI, and TD while minimizing HD and FPR.

\begin{figure}[!b]
    \vspace{-3mm}
    \centering
    \includegraphics[width=3.4in]{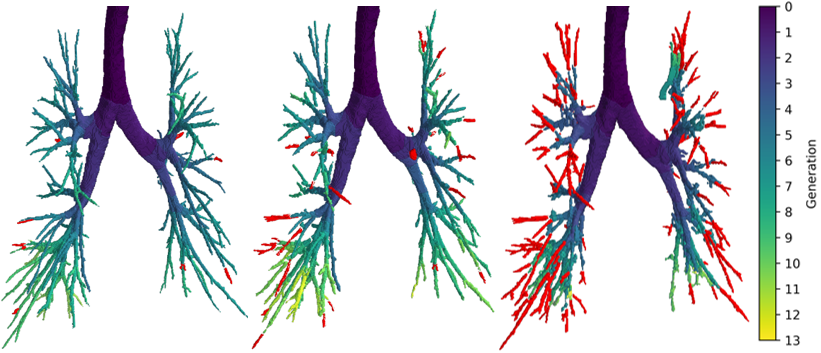}
    
    {\footnotesize
    \hspace*{-0.2in}
    (a) RepAir
    \hspace{0.4in}
    (b) Bronchinet
    \hspace{0.3in}
    (c) NaviAirway
    }
    
    \caption{A 3D rendered view of the segmentation results on a representative image from the ATM'22 test set. \textit{RepAir} is compared with \textit{Bronchinet} and \textit{NaviAirway}. Ground truth airways missed by each model (false negatives) are shown in red, with the predicted airway segmentation color coded by airway generation.}
    \vspace{-5mm}
    \label{fig3}
    \end{figure}

Table~\ref{table1} reports segmentation performance on the ATM’22 and AeroPath test sets. For each dataset and evaluation metric, the highest-performing values are highlighted in bold. From Table~\ref{table1}, we observe that on the ATM’22 dataset, \textit{RepAir} achieves Dice, JI, and TD scores that are higher than \textit{Bronchinet} by 1, 3, and 10 percentage points, respectively, and higher than \textit{NaviAirway} by 13, 21, and 45 percentage points, respectively. Additionally, \textit{RepAir} attains the lowest HD and FPR among all compared methods on the ATM’22 dataset, indicating improved boundary accuracy and reduced false detections. Figure~\ref{fig3} provides a qualitative comparison for a representative test case from the ATM’22 dataset. As shown in Fig.~\ref{fig3}, \textit{RepAir} exhibits the fewest false negatives (color-coded in red), accurately capturing small, high-generation airways that are frequently missed by \textit{Bronchinet} and \textit{NaviAirway}.

We further evaluated all automated methods on the AeroPath dataset, which includes subjects with severe airway pathologies. All models trained exclusively on the ATM’22 training set were directly evaluated without fine-tuning to assess generalization performance. As shown in Table~\ref{table1}, \textit{RepAir} consistently achieves the highest Dice, TPR, JI, and TD scores and the lowest HD compared to \textit{Bronchinet} and \textit{NaviAirway}, confirming its robustness and strong applicability to healthy and diseased lungs.

\section{Conclusion} 

We present \textit{RepAir}, a fully automated airway segmentation framework that enhances airway tree completeness by integrating voxel-level deep learning segmentation with connectivity-aware refinement. \textit{RepAir} employs an nnU-Net-based segmentation model with a connection detection and classification module to address discontinuities in airway trees, reconstructing complete and anatomically consistent airway structures. Experiments on the ATM’22 and AeroPath datasets demonstrate that \textit{RepAir} achieves state-of-the-art performance across both voxel-level and topology-based metrics, outperforming prior 3D U-Net variants such as \textit{Bronchinet} and \textit{NaviAirway}. The framework improves airway connectivity without compromising segmentation accuracy and remains robust in scans with severe disease or structural irregularities. Future work will focus on incorporating connectivity constraints into end-to-end training, extending the framework to multi-task learning for airway wall segmentation, and evaluating its utility in longitudinal disease monitoring.

\bibliographystyle{IEEEtran}

\bibliography{refs}

\end{document}